\documentclass[manuscript,screen,authorversion,acmsmall]{acmart}

\AtBeginDocument{%
  \providecommand\BibTeX{{%
    \normalfont B\kern-0.5em{\scshape i\kern-0.25em b}\kern-0.8em\TeX}}}

\newcommand{\ignore}[1]{}
\usepackage{multirow}

\begin{document}

\title{Predicting the Performance of Multilingual NLP Models}

\author{Anirudh Srinivasan}
\authornote{Work done while the author was at Microsoft Research}
\affiliation{%
  \institution{The University of Texas at Austin}
  \city{Austin}
  \state{Texas}
  \country{USA}
}
\email{anirudhsriniv@gmail.com}

\author{Sunayana Sitaram}
\affiliation{%
  \institution{Microsoft Research}
  \city{Bengaluru}
  \state{Karnataka}
  \country{India}
}
\email{sunayana.sitaram@microsoft.com}

\author{Tanuja Ganu}
\affiliation{%
  \institution{Microsoft Research}
  \city{Bengaluru}
  \state{Karnataka}
  \country{India}
}
\email{tanuja.ganu@microsoft.com}

\author{Sandipan Dandapat}
\affiliation{%
  \institution{Microsoft R\&D}
  \city{Hyderabad}
  \state{Telangana}
  \country{India}
}
\email{sadandap@microsoft.com}

\author{Kalika Bali}
\affiliation{%
  \institution{Microsoft Research}
  \city{Bengaluru}
  \state{Karnataka}
  \country{India}
}
\email{kalikab@microsoft.com}

\author{Monojit Choudhury}
\affiliation{%
  \institution{Microsoft Research}
  \city{Bengaluru}
  \state{Karnataka}
  \country{India}
}
\email{monojitc@microsoft.com}

\renewcommand{\shortauthors}{Srinivasan, et al.}

\begin{abstract}
 Recent advancements in NLP have given us models like mBERT and XLMR that can serve over 100 languages. The languages that these models are evaluated on, however, are very few in number, and it is unlikely that evaluation datasets will cover all the languages that these models support. Potential solutions to the costly problem of dataset creation are to translate datasets to new languages or use template-filling based techniques for creation. This paper proposes an alternate solution for evaluating a model across languages which make use of the existing performance scores of the model on languages that a particular task has test sets for. We train a predictor on these performance scores and use this predictor to predict the model's performance in different evaluation settings. Our results show that our method is effective in filling the gaps in the evaluation for an existing set of languages, but might require additional improvements if we want it to generalize to unseen languages.
\end{abstract}



\maketitle

\ignore{
Introduction
- Lower resource languages, evaluation may not always be possible
- Multilingual models, different work saying this/that feature impacts performance
- Can we define these features and build a model that'll predict performance

Related Work
- Papers predicting performance
- Papers theorizing that this feature impacts performance

Features & Eval method
- Define features, give reasons for why they are chosen
- Define eval methods - E1 and E2

Eval V1
- Single Pivot Eval
- Give Insights on single pivot eval
- Plots - good pivot/bad target etc... - Try to include some actual reasoning here

Eval V2
- Multi Pivot Eval
- Give insights on multi pivot eval
- PLots - lang wise distribution of LOLO error

Eval V2 trends
- Case 1 and Case 2, plus baselines
- Can put plots here
- Baseline insights into difficulty of task
}

\section{Introduction}
Multilingual BERT (mBERT) \cite{devlin-etal-2019-bert} and XLM-Roberta (XLMR) \cite{conneau-etal-2020-unsupervised} are transformer-based encoders that are pretrained on corpora spanning over 100 languages. The pretraining procedure makes use of only unlabelled data (raw text), which is easy to obtain for a large number of languages from various sources including CommonCrawl\footnote{https://commoncrawl.org/} and Wikipedia dumps\footnote{https://dumps.wikimedia.org/}. These models can then be used to solve a variety of downstream tasks by finetuning on a much smaller amount of labelled data for that task.

These models have become very popular due to their ability to transfer downstream task performance across languages in a zero-shot manner, i.e they can be finetuned on labelled data for a task in one language and then be used to solve that task in multiple other languages which do not have any labelled data. This was originally demonstrated on a Natural Language Inference task (XNLI) \cite{conneau-etal-2018-xnli} using mBERT where the model was initially finetuned using only English training data and was then able to solve the task in 14 more languages. Subsequent work \cite{pires-etal-2019-multilingual, wu-dredze-2019-beto} have found similar performance on tasks like Parts-of-Speech tagging (POS), Named Entity Recognition (NER) and Dependency Parsing.

When we look at the popular datasets typically used to evaluate these models, we find that none of them support more than a handful of languages: CoNLL NER \cite{tjong-kim-sang-de-meulder-2003-introduction} - 4 languages, MLQA \cite{lewis-etal-2020-mlqa} - 7 languages, PAWS-X \cite{yang-etal-2019-paws} - 7 languages, TyDiQA-GoldP \cite{clark-etal-2020-tydi} - 9 languages, XQuAD \cite{artetxe-etal-2020-cross} - 11 languages, XNLI \cite{conneau-etal-2018-xnli} - 15. Datasets like UDPOS and WikiAnn are exceptions, spanning a larger number of languages (57 and 89 respectively for the subsets of the dataset(s) that we used), but the number of such exceptions is too few. This disparity can be attributed to 2 reasons. 
First is the fact that there is a disparity in the amount of text resources available in different languages. For example, \citet{joshi-etal-2019-unsung} show that while many Asian and Indian languages have many more speakers than some European languages, they are extremely lacking in the amount of text resources that they have compared to their European counterparts. Many languages spoken in large communities in the world get left behind when it comes to NLP systems built for them.
Second is the fact that even for languages that have text resources, there is a disparity in the amount of labelled data present across languages \cite{joshi-etal-2020-state}. This is likely due to the high cost of annotating datasets as well as the lack of access to language experts or crowd workers for some languages. This leaves us in a situation where multilingual language models (LMs) can be built for these languages, but the LM cannot be evaluated across different tasks in those languages due to the unavailability of labelled data.

The cross-lingual transfer ability of these models make them very useful in scenarios where a downstream task needs to be supported in multiple languages. A single finetuned model can be deployed to serve many languages at once. We however have the problem that we cannot evaluate these models on all the languages that they support, so what we can we do about it? One potential idea is to come up with techniques to automatically or semi-automatically create test sets in new languages. Another way is to use insights from the evaluation on languages and tasks for which labelled data is available and extrapolate these findings to unseen languages.

When it comes to creating test sets in new languages, we could use a couple of techniques to accelerate this process and make it easier. The high quality of current state-of-the-art machine translation (MT) systems \cite{johnson-etal-2017-googles} makes it viable to use MT for translating test sets from existing languages into newer languages. However, labelled datasets require their annotations to be preserved during translation which is difficult to accomplish, particularly for word-level annotations. Another possible method is to use the technique proposed by \citet{ribeiro-etal-2020-beyond}, in which the creation of datasets can be sped up by using masked language models to fill slots in user generated templates. This can also be potentially used along with translation to scale up testing to multiple languages.However, often the creation of initial templates for slot filling is time consuming and requires linguistic expertise.

The other way of solving this does not involve creating test sets, and we look into this in this paper. If we consider the task of XNLI from before, English, the language that mBERT is finetuned on, is considered to be the pivot language, and the 15 languages that the finetuned model is evaluated on are considered as the target languages. More generally, if the task in concern has training data in $p$ pivot languages and test data in $t$ target languages, the model can be finetuned on any mix of data from these $p$ languages and the finetuned model can be evaluated on $t$ languages. If the model is finetuned on just one language, we refer to it just as the pivot language, but if the language is finetuned on a mix of data from multiple pivots, we refer to the mix as a \textit{training configuration}. We should however remember that the model supports many more than the $p$ or $t$ languages that we have data for. 

In this paper looks into the task of predicting the performance of multilingual models on training configurations and languages that we can not easily test for. We define a set of features that characterize the model's performance on a particular language. These features are used to train a predictor on the known (for some configurations) performance scores of the multilingual model to predict performance for the unknown configurations for a given task. The predictor is evaluated via different error metrics simulating different use cases scenarios of the predictor.
\section{Related Work}
mBERT \cite{devlin-etal-2019-bert} and XLMR \cite{conneau-etal-2020-unsupervised} are multilingual transformer models that have been pretrained on text from around 100 languages. These models have been evaluated on tasks like natural language inference, document classification, parts of speech tagging, named entity recognition, dependency parsing \cite{conneau-etal-2018-xnli,pires-etal-2019-multilingual,wu-dredze-2019-beto,wu-dredze-2020-languages} and have shown excellent cross-lingual transfer of performance. These models were found to perform well on tasks involving code-mixed text too \cite{khanuja-etal-2020-gluecos,aguilar-etal-2020-lince}. Given that these models have to support over 100 languages with the limited model capacity they have, some works have found that they are outperformed by monolingual versions, even on some low-resourced languages \cite{devries2019bertje,ortiz-suarez-etal-2020-monolingual,virtanen2019multilingual,pyysalo-etal-2021-wikibert}.

Some of the aforementioned works \cite{pires-etal-2019-multilingual,wu-dredze-2019-beto,wu-dredze-2020-languages} have theorized about the reasons why cross-lingual transfer works and have stated that factors like pretraining data size and vocabulary overlap between languages could affect the transfer performance of a language in these models. \citet{lauscher-etal-2020-zero} study the correlation between transfer performance and factors like syntax, phonology, data size when English is used as the finetuning language. \citet{turc2021revisiting} study the performance of mBERT and mT5 \cite{xue2021mt5} on a wider variety of tasks and look at the impact of using different languages as the finetuning/pivot language.

The task of predicting an NLP model's performance is something that has been looked into in the traditional train-transfer scenario that was popular before pretrained NLP models, with the goal of determining which high-resourced transfer language to use to maximize the performance in a lower-resourced target language. \citet{lin-etal-2019-choosing} develop a set of features based on the overlap between the 2 languages and use that to predict which transfer language would be the best. \citet{xia-etal-2020-predicting,ye-etal-2021-towards} evaluate similar techniques on a wider range of tasks and look into better reliability estimates for prediction. \citet{vu-etal-2020-exploring} meanwhile look at intermediate task finetuning in English, i.e finetuning on a high resourced intermediate task before finetuning on the target task that might have lesser data. They build a set of task embeddings from the transformer model's representations that can be used to predict which intermediate task to finetune on to maximize the performance on the final task. In contrast to these existing works, our work looks specifically into pretrained multilingual transformer models, where a single model can be trained and tested on multiple languages.
\section{Methodology}
In this section, we describe a set of factor that could influence the zero/few-shot performance of a multilingual model. These factors are represented by a set of features for each language. We then briefly describe the regression model used to build the predictor. Furthermore, we go on to describe the evaluation setup where a predictive model is trained to predict the performance of a language given a set of features about the language. We also describe the 2 methods by which the performance of the predictor is evaluated. 

\subsection{Choice of Features}
\label{sec:features_defn}
Most multilingual models are pre-trained on monolingual corpora spanning a large number of languages. For example, mBERT is pre-trained on a Wikipedia corpus spanning 104 languages, while XLMR is pre-trained on a CommonCrawl corpus spanning 100 languages. Some models, such as \citet{lample2019crosslingual, huang-etal-2019-unicoder}, have parallel data added in during pre-training as well. A single shared sub-word vocabulary spanning all the languages is used by these models, leading to overlap between the tokens spanning languages, which could serve as a potential cross-lingual signal \citep{artetxe-etal-2020-call}. Based on these properties of the models, we propose a set of features. The features fall in 2 categories: ones that are dependent only on the target language (Target dependent) and ones that are a function of both the pivot and target language (Pivot-Target dependent). The number of Pivot-Target dependent features that we will have eventually will depend on the number of pivot languages during finetuning.

\subsubsection{Size of pre-training data}
\begin{figure*}[t]
    \centering
    \includegraphics[width=\textwidth]{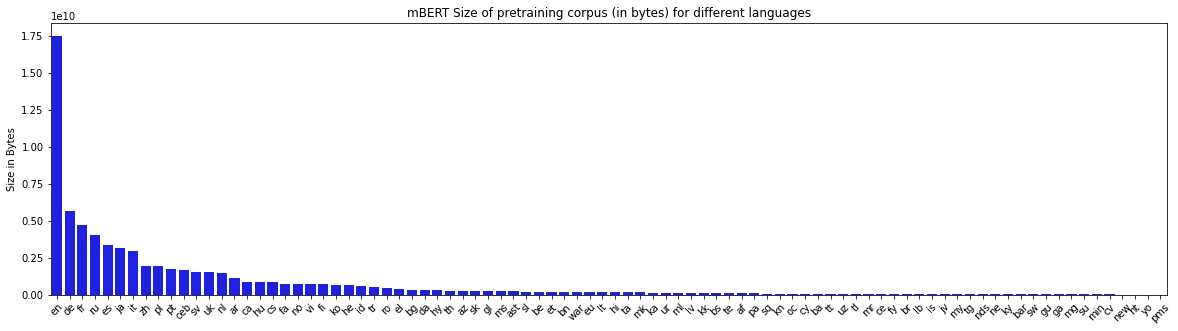}
    \caption{Size of the mBERT's pretraining corpus for different languages. Languages like en have exponentially more data than the next lowest language}
    \label{fig:mbert_pretraining_size}
\end{figure*}

For each language, we take the $log_{10}$ of the size of the corpus used in pre-training. Figure \ref{fig:mbert_pretraining_size} shows that languages like English have exponentially more data than the next lowest language German, so a logarithm was taken to reduce these values to a more linear scale. This metric is independent of the pivot language used for fine-tuning. The metric is denoted in the results as \textbf{Data Size}.

\subsubsection{Typological features}
Typological features capture the structural properties exhibited by languages, and there is evidence that shows that similarity in linguistic typology has an impact on multilingual language modeling \citep{gerz-etal-2018-relation}. One such feature is word order, and \citet{k2020crosslingual} show that it has an impact on cross-lingual performance. \citet{xia-etal-2020-predicting} also find correlations between typology and cross-lingual transfer. Hence, we decide to use typological features as a factor in our predictive model. 

We consult the WALS database \citep{wals} to obtain information about these features. This database comprises of over 2000 languages and 150 typological features, with each language's entry specifying which form (if present) the typological feature takes for the language. The typological feature entries in WALS are sparse, i.e many languages have entries for only a subset of the typological features. If these entries weren't sparse, we could use them as is without any modification or processing. Since this is not the case, we develop a score for each language based on how well represented it's typological features are in the data used for pre-training the model. 

The intuition behind such a score is as follows: suppose, during pre-training, a multilingual model was not exposed to enough text exhibiting a particular feature, it could possibly struggle when it is run on languages having that feature. We quote an example of this from the WALS database. Yoruba uses a \textit{pure vigesimal}\footnote{A Vigesimal number system is a base 20 number system, unlike the more commonly known Decimal number system that is base 10} number system and is one of the few languages among the ones mBERT is pre-trained on that has this number system. Whereas, there are 25+ languages that use the \textit{decimal} number system\footnote{Many of mBERT's languages do not have an entry for their number system in WALS}. Due to this, it is possible that the model is much better at recognizing and handling text with the decimal number system, impacting its performance on a language like Yoruba.

A language having a large number of such features that the model has not seen during pre-training may not perform as well as other languages. Thus, for each language, we define a metric that determines how well represented its typological features are. We use the WALS database to obtain typological features for each language and exclude the features based on phonology, taking the features starting from ``20A'' onward. Since each feature is multi-valued, we binarize the database and derive 541 binary features.

Using the training data sizes for each language, we determine the total amount of training data present for each feature (as the sum of the training data of the languages that exhibit that feature) and rank the features in descending order of data present. Based on these ranked features, we obtain the final metric to estimate how well they are represented:

\begin{align*}
\textrm{Well Rep Feat}(L_i) = \mathop{\huge \text{\LARGE $\mathbb{E}$}}_{k \in \textrm{feat}[i]} \frac{1}{\textrm{rank}(k)}
\end{align*}

where $\textrm{feat}[i]$ are the features that language $L_i$ exhibits and $\textrm{rank}(k)$ is the rank of feature $k$ as per the ordering defined. In essence, \textbf{this metric is the mean reciprocal rank of all the features that language $L_i$ exhibits}. A language with a large number of underrepresented features would have a lower value for this metric as its features would be ranked low, whereas a language with well represented features would have a higher value for this metric. This metric is independent of the pivot language used and is denoted in the results as \textbf{Well Rep Feat}.

\subsubsection{Type overlap with pivot language}
\begin{figure*}[t]
    \centering
    \includegraphics[width=\textwidth]{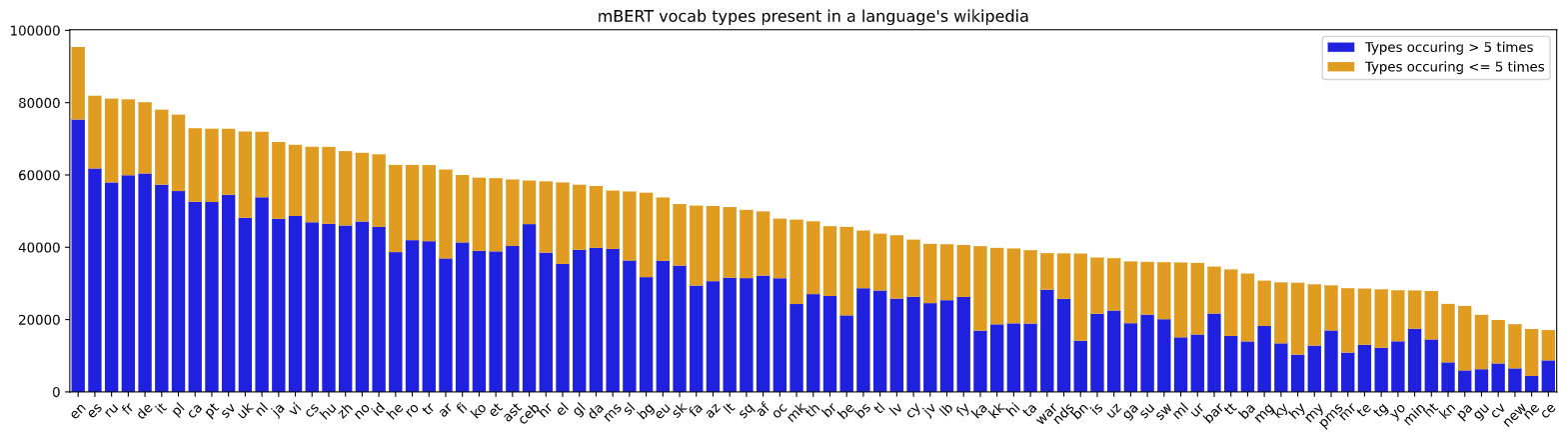}
    \caption{Number of types from the mBERT vocabulary that cover each language. Types highlighted in orange occur $\leq$ 5 times, and contribute to as much as 50\% of the types for some languages}
    \label{fig:mbertvocab}
\end{figure*}

Subword overlap has been used to study the performance of languages in different transfer learning settings and in machine translation \citep{lin-etal-2019-choosing, xia-etal-2020-predicting, nguyen-chiang-2017-transfer}. More recently, the vocabulary overlap between languages has been reported as one of the reasons for mBERT's powerful cross-lingual abilities \citep{pires-etal-2019-multilingual, wu-dredze-2019-beto}. Hence, we wish to incorporate the vocabulary overlap between the target and pivot language into our model. We consider the set of possible subwords (types) that cover a language. For each language $L_i$, we obtain a metric that signifies the overlap between its vocabulary and the vocabulary of the pivot language $L_p$. 

\begin{align*}
\textrm{Type Overlap}(L_i, L_p) = \frac{|T_i| \cap |T_p|}{|T_i| \cup |T_p|}
\end{align*}

where $T_i$ is the set of types that cover language $L_i$. $T_i$ is obtained by using the model's tokenizer. Multilingual Wikipedia dumps were used for this purpose. We discard from $T_i$ types that occur $<=$ 5 times. Figure \ref{fig:mbertvocab} shows that the number of types occurring $<=$ 5 times contribute to as much as 50\% of |$T_i$| for some languages \citep{zipf1949human}. These are unlikely to contribute to the language's performance and hence we discard them before calculating Type Overlap to prevent the metric from being skewed. This metric is dependent on the pivot language used and is denoted in the results as \textbf{Pivot Overlap}.

\subsubsection{Distance from Pivot Language}
For each language $L_i$, we obtain a metric signifying the distance between it and the pivot language $L_t$. For this, we use \texttt{lang2vec} \citep{littell-etal-2017-uriel}, which contains vectors for each language spanning different categories like syntax, phonology etc. The syntax vectors are based on the entries from WALS, SSWL and Ethnologue. The distance between these vectors is used as the distance between the languages. Like the previous metric, this one is dependent on the pivot language used. The metric is denoted in the results as \textbf{Pivot Distance}.

\subsection{Predictive Model}
The aforementioned features are used as inputs for a regression model. Each regression model is trained on performance scores over multiple target languages on a particular task. This is detailed more in the subsequent sections. We use XGBoost \cite{Chen:2016:XST:2939672.2939785} as the regressor. XGBoost has been used in other work that does prediction of the performance of NLP models \cite{lin-etal-2019-choosing, xia-etal-2020-predicting}. We also found that it outperformed a dense netural network when we evaluated on a few dataests. Each regressor is trained with squared error as the loss function, a learning rate of 0.1, max depth of 10 and num estimators of 100. All the input features are converted to a [0, 1] range via min-max normalization. The performance scores used are also converted to a [0, 1] range. The predictor's error is measured using Mean Absolute Error (MAE). This error is also in a [0, 1] range, so an error of 0.04 will  correspond to 4\%. 

\subsection{Evaluating the Predictive Model}
\label{sec:errors}
We evaluate the predictor via two different methods. The difference between these methods is in whether there is any overlap between languages in the data used for training and testing the predictor.  
\begin{itemize}
    \item E1 - In this method, we allow for an overlap between the languages in the train and test set. This represents the scenario where we have a predictor trained on a set of languages and want to use it to predict on these languages but on different training configuration.
    \item E2 - This method is referred to as Leave One Language Out (LOLO). We create a test set containing scores only from a particular language $l$, with the train set not containing any examples (neither as \textit{pivot} or \textit{target}) from language $l$. We measure the error of the predictor in this setup and average it over all the target languages we have. This represents a scenario where a predictor trained on a set of languages and we want to use it on a new language that it has not seen during training.
\end{itemize}
In addition to this, we report a baseline error value for E2. The \textbf{baseline} is a simple mean of the scores using training the XGBoost model in each case.

\section{Experiments: Single Pivot}
\begin{table}[t]
    \centering
    \begin{tabular}{cccc}
        \toprule
        Dataset & \begin{tabular}[c]{@{}l@{}}Pivot\\ Langs\end{tabular} & \begin{tabular}[c]{@{}l@{}}Target\\ Langs\end{tabular} & \begin{tabular}[c]{@{}l@{}}Num Training\\Examples\end{tabular} \\
        \midrule
        XNLI & 15 & 15 & 225 \\
        UDPOS & 29 & 33 & 957 \\
        WikiANN & 40 & 40 & 1600 \\
        \bottomrule
    \end{tabular}
    \caption{Number of Pivot and Target languages in the datasets used for training the predictive model for Section 3. For UDPOS and WikiANN, we restrict the datasets to the languages used by the XTREME benchmark \cite{pmlr-v119-hu20b}}
    \label{tab:dataset_details}
\end{table}

\subsection{Experimental Setup}
We first evaluate our technique in a setup where the multilingual model is finetuned on data from a single pivot language (no mixing of training data from multiple pivot languages). We collect performance scores of mBERT and XLMR Large on existing datasets that are available in multiple pivot and target languages. The details of the datasets used are in Table \ref{tab:dataset_details}. The model is finetuned on each of the $p$ pivot languages separately, and each of these $p$ finetuned models is evaluated on the $t$ target languages. Thus, if a dataset has $p$ pivot languages and $t$ target languages, we obtain $p*t$ performance scores and these are used as the data points for training the predictive model. Since each model is finetuned on a single pivot, we do not need to worry about adding the Pivot-Target dependent features multiple times. We have 4 features (2 Target dependent + 2 Pivot-Target dependent) in total as input for the predictor.

\subsection{Error Computation}
We compute the error in the E1 setup as follows. For each $(pivot,target)$ combination, we take all the $p*t$ data points, leave that particular point out, train a predictor on the remaining $(p*t) - 1$ points and use this predictor to predict the score for the left out point. This gives the estimate for the error for the case when we have some data points for a target language and want to estimate the scores for the missing ones.

The E2 error is computed in the following manner. For each $(pivot,target)$ combination, we take all the $p*t$ data points, leave out the $p$ points corresponding to the target language $target$, train a predictor on the remaining $p*(t-1)$ points and use this predictor to predict the score for the left out point. This gives the estimate for the error for the case when we have no data points for a target language, which is a realistic scenario for most untested languages. The mean baseline for the E2 scenario is computed by taking the mean of the $p*(t-1)$ data points used for training the predictor.

\subsection{Results}

\begin{table}[t]
    \begin{tabular}{llllll}
        \toprule
        Model & Task    & \multicolumn{4}{c}{Feature Importance}                     \\
              &         & Data Size & Well Rep Feat & Pivot Overlap & Pivot Distance \\
        \midrule
              & XNLI    & 0.8317    & 0.0851        & 0.0310        & 0.0521         \\
        mBERT & UDPOS   & 0.1749    & 0.3765        & 0.0987        & 0.3499         \\
              & WikiAnn & 0.2662    & 0.3390        & 0.2236        & 0.1712         \\
        \midrule
              & XNLI    & 0.9218    & 0.0459        & 0.0202        & 0.0121         \\
        XLMR  & UDPOS   & 0.2687    & 0.2073        & 0.1328        & 0.3912         \\
              & WikiAnn & 0.4542    & 0.2344        & 0.2344        & 0.1524         \\
        \bottomrule
    \end{tabular}
    \captionof{table}{Feature Importance values from XGBoost for different input features used}
    \label{tab:feat_imp}
\end{table}

\begin{figure}[t]
    \begin{minipage}{0.49\textwidth}
        \centering
        \begin{tabular}{lllll}
            \toprule
            Model & Dataset & \multicolumn{2}{l}{Predictor Error} & Baseline \\
                  &         & E1               & E2               &          \\
            \midrule
                  & XNLI    & 0.0082           & 0.0279           & 0.0325   \\
            mBERT & UDPOS   & 0.0636           & 0.1015           & 0.1004   \\
                  & WikiAnn & 0.0940           & 0.1554           & 0.1369   \\
            \midrule
                  & XNLI    & 0.0052           & 0.0259           & 0.0298   \\
            XLMR  & UDPOS   & 0.0825           & 0.1149           & 0.1160   \\
                  & WikiAnn & 0.0856           & 0.1285           & 0.1253   \\
            \bottomrule
        \end{tabular}
        \captionof{table}{Two types of predictor error (E1 and E2) and the mean baseline for E2 on different datasets. Lower is better. Predictions/Task Scores are in the range $[0, 1]$, so the same range applies for the error}
        \label{tab:errors}
    \end{minipage}
    \hfill
    \begin{minipage}{0.49\textwidth}
        \centering
        \includegraphics[scale=0.25]{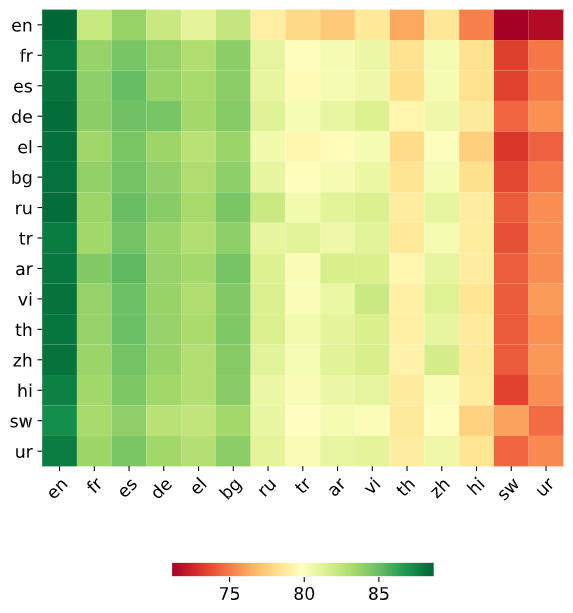}
        \captionof{figure}{XLMR XNLI performance scores for 15 target languages (columns) with 15 pivot languages (rows)}
        \label{fig:xnli_pivots}
    \end{minipage}
\end{figure}

Table \ref{tab:errors} contains the errors for the predictor and the baseline methods. We observe that the E2 error of our predictor is around 2\% on XNLI and 10-15\% on UDPOS and WikiAnn. The errors of both the mean baseline and predictor are very low for XNLI because the range of scores (highest - lowest value observed in training data) is much lower than the other tasks.

The errors in the E1 case are much lower, about half of the E2 error in many cases. The difference between the E1 and E2 methods is that in the E1 method, the model has been exposed to data points with the same language as the one on which error has been calculated. This suggests that the predictor's ability to generalize to unseen languages isn't the best across tasks. However, we have to keep in mind that the E1 and E2 settings are 2 extreme ends of a spectrum (w.r.t the number of examples used from a particular target language). The predictor's performance on a new language could potentially be improved by adding just a few examples from that language.

\subsection{Observations}
\subsubsection{Trends in Feature Importance Values}
Table \ref{tab:feat_imp} contains the feature importance values returned by the XGBoost predictor used for the different tasks. We observe a clear difference in which features are important between XNLI, a semantic task, and UDPOS/WikiAnn, which are more syntactic tasks. For syntactic tasks, the predictor relies more on the features that are based on typology and the relation/overlap between languages. For the semantic task, the predictor is relying mainly on the pretraining data size. This could be a more general indication of the performance of multilingual models on different tasks, possibly suggesting that using more pretraining data is likely to a help a language more in semantic tasks, while syntactic tasks are likely to benefit not by this, but rather by finetuning on related languages.

\subsubsection{Performance of Target Languages on Different Pivots}

\begin{figure}[t]
    \centering
    \includegraphics[scale=0.3]{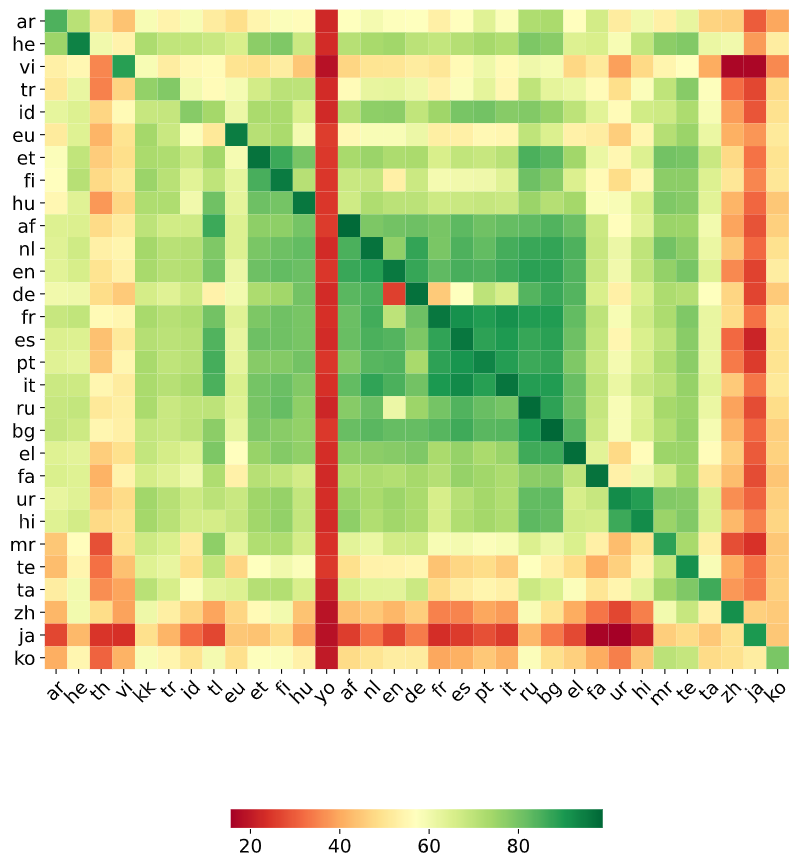}
    \caption{XLMR UDPOS performance scores for 33 target languages (columns) with 29 pivot languages (rows)}
    \label{fig:udpos_pivots}
\end{figure}

\begin{figure}[t]
    \centering
    \includegraphics[scale=0.35]{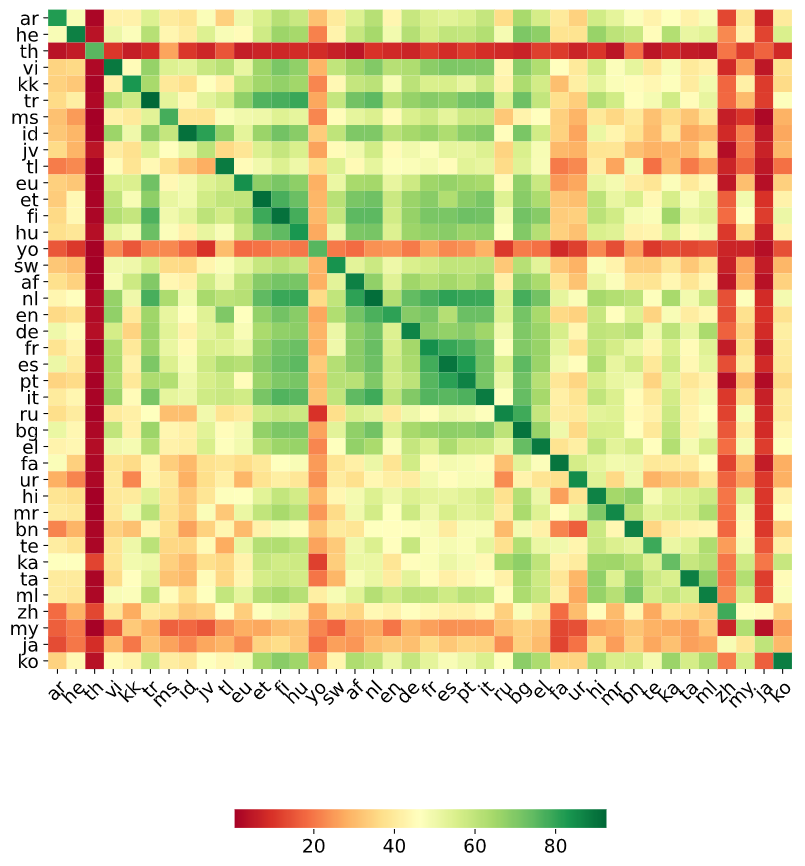}
    \caption{XLMR WikiAnn performance scores for 40 target languages (columns) with 40 pivot languages (rows)}
    \label{fig:wikiann_pivots}
\end{figure}

In the process of collecting data points to train the predictor, we end up with the performance scores of $p$ finetuned models on $t$ target languages for each task. These performance scores, on XLMR, are visualized in the form of heatmaps (Figures \ref{fig:xnli_pivots}, \ref{fig:udpos_pivots}, \ref{fig:wikiann_pivots}) and these heatmaps give us some insights into the performance of the multilingual models themselves.

Figure \ref{fig:xnli_pivots} shows the distribution of scores on XNLI. The languages along the X-axis and Y-axis here are sorted by the amount of resources available in them. The scores on a particular target do not vary significantly depending on the pivot. English does well irrespective of which pivot language is used, and the performance gradually decreases as we move towards targets on the right.

Figures \ref{fig:udpos_pivots} and \ref{fig:wikiann_pivots} show similar heatmaps for UDPOS and WikiAnn. In these plots, the languages along the X-axis and Y-axis are grouped based on language family and arranged as per this grouping. With this kind of an ordering, we are able to see many cases of transfer happening within language families. This is present in the form of green boxes around the diagonal in many instances. We can see this clearly for the languages from the Indo-European family(from Afrikans(af) to Bulgarian(bg)), and to a lesser extent among the Indo-Aryan/Dravidian languages (from Urdu(ur) to Malayalam(ml)).

We also see instances of rows or columns that are completely red. These are cases where transfer across languages does not happen at all. The \textit{red rows} are what we refer to as \textbf{Bad Pivots}. This is the case for Japanese(ja), Chinese(zh), Korean(ko) for UDPOS and Thai(th), Yoruba(yo), Burmese(my), Japanese(ja) on WikiAnn. Finetuning with data on these languages does not seem to help any language other than itself. The \textit{red columns} are what we refer to as \textbf{Bad Targets}. This is the case for Yoruba(yo), Japanese(ja) for UDPOS and Thai(th), Yoruba(yo), Chinese(zh), Japanese(ja) for WikiAnn. Finetuning on languages other than themselves do not help these languages. There is also the case of German(de) in UDPOS, where finetuning on it produces very bad performance on en alone, a language very similar to it. 

We see that there is a good overlap between the languages in Bad Pivots and Bad Targets. This could suggest that the reason for these phenomenon could be that the multilingual model isn't able to support these languages well. The bad performance of the (de, en) pair is a strange one as en is the only language on which the de model does not do well. This could possibly due to some difference in the annotation done for both datasets.

\section{Experiments: Multi Pivot}
\label{sec:multi_pivot}
\begin{table}[t]
    \centering
    \begin{tabular}{cccc}
        \toprule
        Dataset & \begin{tabular}[c]{@{}l@{}}Training Data\\ Combinations\end{tabular} & \begin{tabular}[c]{@{}l@{}}Target\\ Langs\end{tabular} & \begin{tabular}[c]{@{}l@{}}Num Training\\Examples \end{tabular} \\
        \midrule
        XNLI & 256 & 15 & 3840 \\
        UDPOS & 1664 & 57 & 94848 \\
        WikiANN & 1592 & 89 & 141688 \\
        \bottomrule
    \end{tabular}
    \caption{Number of Input Data combinations (training runs) and Target languages in the datasets used for training the predictive model for Section 4. For UDPOS and WikiANN, we used versions of these datasets with more languages than the previous section (which was limited to the XTREME benchamark's languages)}
    \label{tab:dataset_details_v2}
\end{table}

\subsection{Experimental Setup}
We now evaluate our technique in a setup where the multilingual model can be finetuned on a mix of data from $p$ different pivot languages. We finetune XLMR Large on existing datasets in this manner, finetuning on $m$ different combinations of training data across the $p$ languages. Each of these $m$ models are then evaluated on $t$ target languages and we end up with $m*t$ examples in total for the predictor. The details about the datasets are in Table \ref{tab:dataset_details_v2}.

Since we have multiple pivot languages, we need to change the set of input features to the model. Each training configuration has a mix of data from the $p$ pivot languages. We represent this via a $p$ dimensional vector (referred to as the input data vector or finetuning data size), with element $i$ of the vector representing the amount of data used from pivot language $i$ during finetuning. Also, since we have not 1 but $p$ pivot languages, we will need to increase the number of Pivot-Target dependent features from 1 to $p$ to capture the overlap between each pivot and the target. We summarize the set of input features in Table \ref{tab:input_features_v2}.

\begin{table}[t]
\begin{tabular}{@{}lll@{}}
\toprule
Feature                   & Dependency & Number of Features \\ \midrule
Pretraining Data Size     & $t$        & $1$                \\
Well Represented Features & $t$        & $1$                \\
Pivot Overlap             & $(p,t)$    & $p$                \\
Pivot Distance            & $(p,t)$    & $p$                \\
Finetuning Data Size      & $p$        & $p$                \\ \bottomrule
\end{tabular}
\caption{$t$-target, $p$-pivot. Number of features used to train predictor in Multi Pivot setting}
\label{tab:input_features_v2}
\end{table}

Given that we have more than 50 pivot languages in some datasets, we experiment with different methods of adding in the Pivot-Target dependent features, so as to prevent the model from having a very large input space
\begin{enumerate}
    \item None - None of these features are added
    \item All - $2p$ features are added like described above
    \item Data-Only - $2p$ features are added like above, but we then 0 out the values for languages that have a 0 in their training configuration
\end{enumerate}
The total number of input features is either $2+p$ or $2+3p$ depending on the method used

\subsection{Error Computation}
The E1 error (cf. Section \ref{sec:errors}) is computed as follows. We divide the data into 5 folds and compute the error in a 5-fold cross-validation manner (train on 4 folds, test on the 5th fold). The E1 error in the previous section was computed in a similar manner, the only difference being that it used a $n$ fold cross-validation instead of the 5-fold we are using here ($n$ is the total number of training examples). We reduce $n$ to 5 here as $n$ is large and the error computation will be very computationally expensive.
 
The E2 error is computed for each target language in the following manner and is then averaged over all the languages. We sample 1000 train and 100 test points from the data. The test data points are all from the target language and the train data points do not contain any data points from the target language. In addition to this, we ensure that the train data points have the input data vector's value for the target language to be 0, i.e no data from that language was in the mix of data used to finetune the model. This models the scenario where we have training (pivot) or test (target) data for the language. The mean baseline for the E2 scenario is similarly computed by taking the mean of the 1000 data points used for training the predictor.

\subsection{Results}
\begin{table*}[t]
\centering
\begin{tabular}{lllll}
\toprule
Model/Dataset            & Pivot Feats & E1     & E2     & Baseline \\
\midrule
\multirow{3}{*}{UDPOS}   & None         & 0.0040 & 0.0783 & \multirow{3}{*}{0.0786}  \\
                         & All          & 0.0039 & 0.0686 &   \\
                         & Data-Only    & 0.0067 & 0.0736 &   \\
\midrule
\multirow{3}{*}{WikiAnn} & None         & 0.0061 & 0.0993 & \multirow{3}{*}{0.0953}  \\
                         & All          & 0.0063 & 0.0915 &   \\
                         & Data-Only    & 0.0085 & 0.0743 &   \\
\midrule
\multirow{3}{*}{XNLI}    & None         & 0.0129 & 0.0286 & \multirow{3}{*}{0.0541}  \\
                         & All          & 0.0105 & 0.0244 &   \\
                         & Data-Only    & 0.0149 & 0.0307 &   \\
\bottomrule
\end{tabular}
\caption{2 types of predictor error and one baseline error for different datasets on XLMR in the mixed language finetuning setup. Lower is better. E1 - Similar to Leave One Out, E2 - Similar to Leave One Target Out. Predictions/Task Scores are in the range $[0, 1]$, so the same range applies for the error}
\label{tab:v2_results}
\end{table*}

Table \ref{tab:v2_results} contains the errors for the predictor in the new evaluation setup. We can see that the E1 errors are extremely low, while the E2 errors are reasonably high, particularly for UDPOS and WikiAnn. The E2 error is still lower than the mean baseline in most cases, so using the predictor brings in an improvement. The E1 errors being low imply that the predictor is able to predict well on different combinations of finetuning data for languages that it has seen during training. The higher E2 errors indicate that the ability of the predictor to generalize to unseen languages relatively harder.

We varied the set of Type 2 features (Pivot Feats in Table \ref{tab:v2_results}), to see if adding them brought any improvement to the predictor. The performance of each method varies and there does not seem to be a single best option. For both the All and Data-Only option, we end up adding a large number of input features to the model (30 for XNLI, 50+ for UDPOS and WikiAnn). Adding these in does not seem to affect the predictor's performance, so we conduct the rest of the experiments using \textit{None} option.

\begin{figure}[t]
    \centering
    \includegraphics[width=\textwidth]{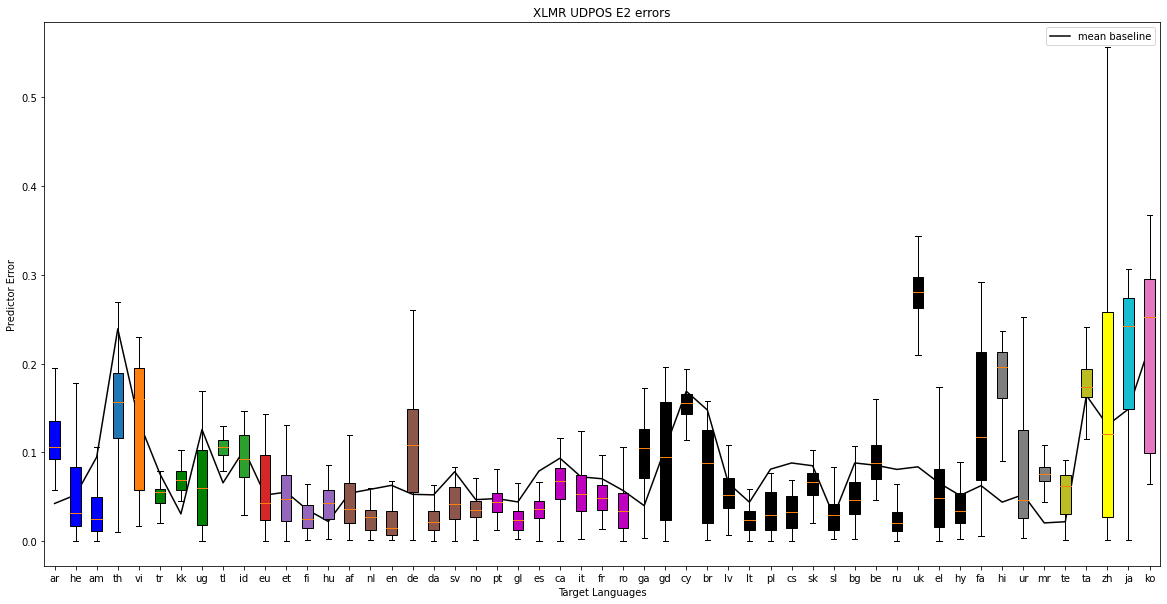}
    \caption{Distribution of E2 Errors across different languages along with the error of the mean baseline method}
    \label{fig:udpos_e2_errors}
\end{figure}

\textbf{Distribution of Errors Across Languages}: Since the E2 errors were higher overall, we take the E2 errors for each language (without averaging over all languages) and plot the distribution of this across different languages in Figure \ref{fig:udpos_e2_errors}. This figure is for the predictor trained on the UDPOS dataset. The error of the baseline error (mean baseline for E2) is depicted by the black line. The languages are coloured and ordered by language family. We can see that the E2 errors vary significantly across languages. For many languages, the errors are under 0.05 (5\%), but for others, the distribution is very wide. The rightmost 3 languages, zh, ja and ko show some of the widest ranges in error.

\section{Relative Evaluation of Predictor}
We have observed so far that the predictor's E1 performance is very good, but the errors in the E2 scenario are not as low as the E1 scenario. To get a better insight into the functioning of the predictor, we propose a new method of evaluation that does not look at the magnitude of the predictor's errors, but rather at whether the relative trends in the predictions match the test data. We construct 2 scenarios for testing the predictor. The test set for each scenario contains a set of pairs of data points. Predictions are computed for each pair in the test set and the relation between them is obtained (either $>$, $<$ or $=$). This is then compared to the relation between the gold scores. We compute and report the accuracy of the relation matching over pairs of data points sampled from the test set.

\subsection{Evaluation Setup}
We use the same data collected for evaluation in Section \ref{sec:multi_pivot}. This data contains $m$ training configs and the performance score for $t$ different languages for each config. This is split into a train and test set, and a predictor is trained on the train set. Suppose $L$ is a target language and $a$ is training configuration. For the data points in the test set, $score(L, a)$ represents the actual score for this combination in the data and $prediction(L, a)$ represents the models prediction for this data point. Given these definitions, the 2 evaluation scenarios are described below:

\textbf{Case 1} (Keep target language same, vary training config): For a target language $T$ and 2 training configs $a$ and $b$ we check if the relation between $score(T, a)$ and $score(T, b)$ is the same as the relation between $prediction(T, a)$ and $prediction(T, b)$

\textbf{Case 2} (Keep training config same, vary target language): For a target languages $T1$ and $T2$ and a training config $a$, we check if the relation between $score(T1, a)$ and $score(T2, a)$ is the same as the relation between $prediction(T1, a)$ and $prediction(T2, a)$

We report the accuracy for both the Case 1 and Case 2 evaluation methods. In addition to this, we implement a naive baseline for each method and report it's accuracy. The baseline methods learn an ordering of the data points based on the target language or the training config and make a prediction based on the ordering. These baselines should give an estimate of how difficult each task is.

\textbf{Case 1 Baseline}: Since we vary the training config while making the comparison, the baseline is an ordering of scores is based on the sum of data across pivot languages in the training config.

\textbf{Case 2 Baseline}: Since we vary the target language while making the comparison, we learn a single score for each target language (as a mean of all it's examples) and do the ordering based on this. In the case where a target language was not seen during training, we output the mean of all the learnt language scores as the prediction.

We simulate both the E1 and E2 evaluation scenarios just like before, by computing the train/test sets in appropriate ways.

\textbf{E1}: For the E1 scenario, we create a 80/20 train/test split of the data. On the test split, we sample data points appropriately for Case 1 and Case 2 by taking points that have a common target language/training config (for Case 1/Case 2, respectively).

\textbf{E2}: For the E2 scenario, we sample 1000 train and 100 test points from the data just like before, ensuring that the test language overlap between train and test is not present. For Case 1, since the target language $T$ is kept constant and only the training configs $a$ and $b$ are varied, all the test points can be from one language. For Case 2, since the target language is varied (as $T1$ or $T2$), the test split is created such that all the data points for $T1$ are from the test language and all the data points for $T2$ are from all other possible languages.




\begin{table}[t]
    \begin{tabular}{llllll}
        \toprule
        Dataset                  & Accuracy  & \multicolumn{2}{l}{Case 1} & \multicolumn{2}{l}{Case 2} \\
                                 &           & E1           & E2          & E1           & E2          \\
        \midrule
        \multirow{2}{*}{UDPOS}   & Predictor & 92.60        & 92.56       & 98.28        & 96.48       \\
                                 & Baseline  & 49.22        & 48.89       & 82.73        & 46.91       \\
        \midrule
        \multirow{2}{*}{WikiAnn} & Predictor & 90.10        & 90.08       & 98.05        & 95.52       \\
                                 & Baseline  & 52.83        & 52.93       & 89.25        & 47.97       \\
        \midrule
        \multirow{2}{*}{XNLI}    & Predictor & 95.71        & 95.78       & 99.02        & 98.26       \\
                                 & Baseline  & 63.48        & 64.06       & 71.04        & 40.10       \\
        \bottomrule
    \end{tabular}
    \caption{Accuracy of the Predictor and Baseline methods on the relative evaluation method. Higher is better.}
\end{table}

\subsection{Results}
\subsubsection{Case 1}
We observe that the predictor's accuracies are extremely high in both the E1 and E2 cases, and that the baseline accuracies are low. The baseline here is just based on the sum of the training data sizes in the pivots, and the low performance numbers indicate that an ordering just based on this is not enough to predict the relative trends in the data. Since we vary the training configuration for Case 1, while keeping the target language same, the high predictor accuracies indicate that the predictor has modelled which pivot languages are important in affecting the performance score.

\subsubsection{Case 2}
The predictor's accuracies are high again, both in the E1 and E2 cases, but we can see that the baseline accuracy is high in the E1 case. The baseline method learns a single value for each target language. It's accuracy being high indicates that this task is a much simpler task and that the model doing well here could just mean that it's learning a similar mapping. The baseline accuracy in the E2 case however is extremely low. Since the target language being tested here is something that was not present during training, the baseline cannot learn a single value for it and it struggles. The predictor however does well here, and this gives us the indication that it's able to generalize to a new language, atleast in terms of the relative values that it predicts.

\subsection{Predictor's Absolute vs Relative Performance}
Although the absolute errors in the E2 case are high, the predictor does well when we just look at the relative trends in the predictions. It is able to generalize to a new language in this manner as well. For an input example in a new language, the predictor has learnt which direction to push the prediction towards based on the features, but is unable to predict the absolute value. There are potential uses for such a predictor where we do not want it to predict the absolute performance value, but just want it to compare 2 training configs or 2 target languages. One such potential use could be to look at which pivot languages to add training data in, to maximize performance. By iterating through the space of training data configs, we only need to make comparison based operations to pick which language to add data in. Based on these results, the predictor would be expected to do well in such a scenario.
\section{Conclusion}
Evaluating large multilingual models on all possible tasks and languages is challenging due to the unavailability of labelled data in most languages. However, it is crucial to be able to evaluate LMs across all languages they serve. In this work, we propose using a model to predict the performance of a multilingual model on different (pivot, target) combinations to serve as a proxy for creating test sets for individual languages and tasks. The model uses a set of features about the pivot and target languages, including typological features and ones that capture language similarity. These are fed to a regression model to predict scores for unseen languages.

We observe that the feature importances returned by our predictor give insights into what factors are more important in predicting for different tasks, with semantic tasks relying more on pretraining and semantic tasks relying more on the typology and subword vocabulary. We compare the predictive model with baselines that take into account the average performance on a pivot language and find that the predictive model only performs well when predicting on a language for which some (pivot, target) data points are available. We also proposed an evaluation method that only looks at the relative values between the predictions. Although we find that the predictor does much better in this form of evaluation, the results indicate that the ability of the predictive model to generalize to unseen languages is still limited, suggesting that improvements are needed to be able to replace the creation of test datasets for new languages.

\bibliographystyle{ACM-Reference-Format}
\bibliography{sample-bib}


\begin{thebibliography}{39}


\ifx \showCODEN    \undefined \def \showCODEN     #1{\unskip}     \fi
\ifx \showDOI      \undefined \def \showDOI       #1{#1}\fi
\ifx \showISBNx    \undefined \def \showISBNx     #1{\unskip}     \fi
\ifx \showISBNxiii \undefined \def \showISBNxiii  #1{\unskip}     \fi
\ifx \showISSN     \undefined \def \showISSN      #1{\unskip}     \fi
\ifx \showLCCN     \undefined \def \showLCCN      #1{\unskip}     \fi
\ifx \shownote     \undefined \def \shownote      #1{#1}          \fi
\ifx \showarticletitle \undefined \def \showarticletitle #1{#1}   \fi
\ifx \showURL      \undefined \def \showURL       {\relax}        \fi
\providecommand\bibfield[2]{#2}
\providecommand\bibinfo[2]{#2}
\providecommand\natexlab[1]{#1}
\providecommand\showeprint[2][]{arXiv:#2}

\bibitem[\protect\citeauthoryear{Aguilar, Kar, and Solorio}{Aguilar
  et~al\mbox{.}}{2020}]%
        {aguilar-etal-2020-lince}
\bibfield{author}{\bibinfo{person}{Gustavo Aguilar}, \bibinfo{person}{Sudipta
  Kar}, {and} \bibinfo{person}{Thamar Solorio}.}
  \bibinfo{year}{2020}\natexlab{}.
\newblock \showarticletitle{{L}in{CE}: A Centralized Benchmark for Linguistic
  Code-switching Evaluation}. In \bibinfo{booktitle}{\emph{Proceedings of the
  12th Language Resources and Evaluation Conference}}.
  \bibinfo{publisher}{European Language Resources Association},
  \bibinfo{address}{Marseille, France}, \bibinfo{pages}{1803--1813}.
\newblock
\urldef\tempurl%
\url{https://aclanthology.org/2020.lrec-1.223}
\showURL{%
\tempurl}


\bibitem[\protect\citeauthoryear{Artetxe, Ruder, and Yogatama}{Artetxe
  et~al\mbox{.}}{2020a}]%
        {artetxe-etal-2020-cross}
\bibfield{author}{\bibinfo{person}{Mikel Artetxe}, \bibinfo{person}{Sebastian
  Ruder}, {and} \bibinfo{person}{Dani Yogatama}.}
  \bibinfo{year}{2020}\natexlab{a}.
\newblock \showarticletitle{On the Cross-lingual Transferability of Monolingual
  Representations}. In \bibinfo{booktitle}{\emph{Proceedings of the 58th Annual
  Meeting of the Association for Computational Linguistics}}.
  \bibinfo{publisher}{Association for Computational Linguistics},
  \bibinfo{address}{Online}, \bibinfo{pages}{4623--4637}.
\newblock
\urldef\tempurl%
\url{https://doi.org/10.18653/v1/2020.acl-main.421}
\showDOI{\tempurl}


\bibitem[\protect\citeauthoryear{Artetxe, Ruder, Yogatama, Labaka, and
  Agirre}{Artetxe et~al\mbox{.}}{2020b}]%
        {artetxe-etal-2020-call}
\bibfield{author}{\bibinfo{person}{Mikel Artetxe}, \bibinfo{person}{Sebastian
  Ruder}, \bibinfo{person}{Dani Yogatama}, \bibinfo{person}{Gorka Labaka},
  {and} \bibinfo{person}{Eneko Agirre}.} \bibinfo{year}{2020}\natexlab{b}.
\newblock \showarticletitle{A Call for More Rigor in Unsupervised Cross-lingual
  Learning}. In \bibinfo{booktitle}{\emph{Proceedings of the 58th Annual
  Meeting of the Association for Computational Linguistics}}.
  \bibinfo{publisher}{Association for Computational Linguistics},
  \bibinfo{address}{Online}, \bibinfo{pages}{7375--7388}.
\newblock
\urldef\tempurl%
\url{https://doi.org/10.18653/v1/2020.acl-main.658}
\showDOI{\tempurl}


\bibitem[\protect\citeauthoryear{Chen and Guestrin}{Chen and Guestrin}{2016}]%
        {Chen:2016:XST:2939672.2939785}
\bibfield{author}{\bibinfo{person}{Tianqi Chen} {and} \bibinfo{person}{Carlos
  Guestrin}.} \bibinfo{year}{2016}\natexlab{}.
\newblock \showarticletitle{{XGBoost}: A Scalable Tree Boosting System}. In
  \bibinfo{booktitle}{\emph{Proceedings of the 22nd ACM SIGKDD International
  Conference on Knowledge Discovery and Data Mining}} (San Francisco,
  California, USA) \emph{(\bibinfo{series}{KDD '16})}.
  \bibinfo{publisher}{ACM}, \bibinfo{address}{New York, NY, USA},
  \bibinfo{pages}{785--794}.
\newblock
\showISBNx{978-1-4503-4232-2}
\urldef\tempurl%
\url{https://doi.org/10.1145/2939672.2939785}
\showDOI{\tempurl}


\bibitem[\protect\citeauthoryear{Clark, Choi, Collins, Garrette, Kwiatkowski,
  Nikolaev, and Palomaki}{Clark et~al\mbox{.}}{2020}]%
        {clark-etal-2020-tydi}
\bibfield{author}{\bibinfo{person}{Jonathan~H. Clark}, \bibinfo{person}{Eunsol
  Choi}, \bibinfo{person}{Michael Collins}, \bibinfo{person}{Dan Garrette},
  \bibinfo{person}{Tom Kwiatkowski}, \bibinfo{person}{Vitaly Nikolaev}, {and}
  \bibinfo{person}{Jennimaria Palomaki}.} \bibinfo{year}{2020}\natexlab{}.
\newblock \showarticletitle{{T}y{D}i {QA}: A Benchmark for Information-Seeking
  Question Answering in Typologically Diverse Languages}.
\newblock \bibinfo{journal}{\emph{Transactions of the Association for
  Computational Linguistics}}  \bibinfo{volume}{8} (\bibinfo{year}{2020}),
  \bibinfo{pages}{454--470}.
\newblock
\urldef\tempurl%
\url{https://doi.org/10.1162/tacl_a_00317}
\showDOI{\tempurl}


\bibitem[\protect\citeauthoryear{Conneau, Khandelwal, Goyal, Chaudhary, Wenzek,
  Guzm{\'a}n, Grave, Ott, Zettlemoyer, and Stoyanov}{Conneau
  et~al\mbox{.}}{2020}]%
        {conneau-etal-2020-unsupervised}
\bibfield{author}{\bibinfo{person}{Alexis Conneau}, \bibinfo{person}{Kartikay
  Khandelwal}, \bibinfo{person}{Naman Goyal}, \bibinfo{person}{Vishrav
  Chaudhary}, \bibinfo{person}{Guillaume Wenzek}, \bibinfo{person}{Francisco
  Guzm{\'a}n}, \bibinfo{person}{Edouard Grave}, \bibinfo{person}{Myle Ott},
  \bibinfo{person}{Luke Zettlemoyer}, {and} \bibinfo{person}{Veselin
  Stoyanov}.} \bibinfo{year}{2020}\natexlab{}.
\newblock \showarticletitle{Unsupervised Cross-lingual Representation Learning
  at Scale}. In \bibinfo{booktitle}{\emph{Proceedings of the 58th Annual
  Meeting of the Association for Computational Linguistics}}.
  \bibinfo{publisher}{Association for Computational Linguistics},
  \bibinfo{address}{Online}, \bibinfo{pages}{8440--8451}.
\newblock
\urldef\tempurl%
\url{https://doi.org/10.18653/v1/2020.acl-main.747}
\showDOI{\tempurl}


\bibitem[\protect\citeauthoryear{Conneau, Rinott, Lample, Williams, Bowman,
  Schwenk, and Stoyanov}{Conneau et~al\mbox{.}}{2018}]%
        {conneau-etal-2018-xnli}
\bibfield{author}{\bibinfo{person}{Alexis Conneau}, \bibinfo{person}{Ruty
  Rinott}, \bibinfo{person}{Guillaume Lample}, \bibinfo{person}{Adina
  Williams}, \bibinfo{person}{Samuel Bowman}, \bibinfo{person}{Holger Schwenk},
  {and} \bibinfo{person}{Veselin Stoyanov}.} \bibinfo{year}{2018}\natexlab{}.
\newblock \showarticletitle{{XNLI}: Evaluating Cross-lingual Sentence
  Representations}. In \bibinfo{booktitle}{\emph{Proceedings of the 2018
  Conference on Empirical Methods in Natural Language Processing}}.
  \bibinfo{publisher}{Association for Computational Linguistics},
  \bibinfo{address}{Brussels, Belgium}, \bibinfo{pages}{2475--2485}.
\newblock
\urldef\tempurl%
\url{https://doi.org/10.18653/v1/D18-1269}
\showDOI{\tempurl}


\bibitem[\protect\citeauthoryear{de~Vries, van Cranenburgh, Bisazza, Caselli,
  van Noord, and Nissim}{de~Vries et~al\mbox{.}}{2019}]%
        {devries2019bertje}
\bibfield{author}{\bibinfo{person}{Wietse de Vries}, \bibinfo{person}{Andreas
  van Cranenburgh}, \bibinfo{person}{Arianna Bisazza}, \bibinfo{person}{Tommaso
  Caselli}, \bibinfo{person}{Gertjan van Noord}, {and} \bibinfo{person}{Malvina
  Nissim}.} \bibinfo{year}{2019}\natexlab{}.
\newblock \bibinfo{title}{BERTje: A Dutch BERT Model}.
\newblock
\newblock
\showeprint[arxiv]{1912.09582}~[cs.CL]


\bibitem[\protect\citeauthoryear{Devlin, Chang, Lee, and Toutanova}{Devlin
  et~al\mbox{.}}{2019}]%
        {devlin-etal-2019-bert}
\bibfield{author}{\bibinfo{person}{Jacob Devlin}, \bibinfo{person}{Ming-Wei
  Chang}, \bibinfo{person}{Kenton Lee}, {and} \bibinfo{person}{Kristina
  Toutanova}.} \bibinfo{year}{2019}\natexlab{}.
\newblock \showarticletitle{{BERT}: Pre-training of Deep Bidirectional
  Transformers for Language Understanding}. In
  \bibinfo{booktitle}{\emph{Proceedings of the 2019 Conference of the North
  {A}merican Chapter of the Association for Computational Linguistics: Human
  Language Technologies, Volume 1 (Long and Short Papers)}}.
  \bibinfo{publisher}{Association for Computational Linguistics},
  \bibinfo{address}{Minneapolis, Minnesota}, \bibinfo{pages}{4171--4186}.
\newblock
\urldef\tempurl%
\url{https://doi.org/10.18653/v1/N19-1423}
\showDOI{\tempurl}


\bibitem[\protect\citeauthoryear{Dryer and Haspelmath}{Dryer and
  Haspelmath}{2013}]%
        {wals}
\bibfield{editor}{\bibinfo{person}{Matthew~S. Dryer} {and}
  \bibinfo{person}{Martin Haspelmath}} (Eds.). \bibinfo{year}{2013}\natexlab{}.
\newblock \bibinfo{booktitle}{\emph{WALS Online}}.
\newblock \bibinfo{publisher}{Max Planck Institute for Evolutionary
  Anthropology}, \bibinfo{address}{Leipzig}.
\newblock
\urldef\tempurl%
\url{https://wals.info/}
\showURL{%
\tempurl}


\bibitem[\protect\citeauthoryear{Gerz, Vuli{\'c}, Ponti, Reichart, and
  Korhonen}{Gerz et~al\mbox{.}}{2018}]%
        {gerz-etal-2018-relation}
\bibfield{author}{\bibinfo{person}{Daniela Gerz}, \bibinfo{person}{Ivan
  Vuli{\'c}}, \bibinfo{person}{Edoardo~Maria Ponti}, \bibinfo{person}{Roi
  Reichart}, {and} \bibinfo{person}{Anna Korhonen}.}
  \bibinfo{year}{2018}\natexlab{}.
\newblock \showarticletitle{On the Relation between Linguistic Typology and
  (Limitations of) Multilingual Language Modeling}. In
  \bibinfo{booktitle}{\emph{Proceedings of the 2018 Conference on Empirical
  Methods in Natural Language Processing}}. \bibinfo{publisher}{Association for
  Computational Linguistics}, \bibinfo{address}{Brussels, Belgium},
  \bibinfo{pages}{316--327}.
\newblock
\urldef\tempurl%
\url{https://doi.org/10.18653/v1/D18-1029}
\showDOI{\tempurl}


\bibitem[\protect\citeauthoryear{Hu, Ruder, Siddhant, Neubig, Firat, and
  Johnson}{Hu et~al\mbox{.}}{2020}]%
        {pmlr-v119-hu20b}
\bibfield{author}{\bibinfo{person}{Junjie Hu}, \bibinfo{person}{Sebastian
  Ruder}, \bibinfo{person}{Aditya Siddhant}, \bibinfo{person}{Graham Neubig},
  \bibinfo{person}{Orhan Firat}, {and} \bibinfo{person}{Melvin Johnson}.}
  \bibinfo{year}{2020}\natexlab{}.
\newblock \showarticletitle{{XTREME}: A Massively Multilingual Multi-task
  Benchmark for Evaluating Cross-lingual Generalisation}. In
  \bibinfo{booktitle}{\emph{Proceedings of the 37th International Conference on
  Machine Learning}} \emph{(\bibinfo{series}{Proceedings of Machine Learning
  Research}, Vol.~\bibinfo{volume}{119})},
  \bibfield{editor}{\bibinfo{person}{Hal~Daumé III} {and}
  \bibinfo{person}{Aarti Singh}} (Eds.). \bibinfo{publisher}{PMLR},
  \bibinfo{pages}{4411--4421}.
\newblock
\urldef\tempurl%
\url{http://proceedings.mlr.press/v119/hu20b.html}
\showURL{%
\tempurl}


\bibitem[\protect\citeauthoryear{Huang, Liang, Duan, Gong, Shou, Jiang, and
  Zhou}{Huang et~al\mbox{.}}{2019}]%
        {huang-etal-2019-unicoder}
\bibfield{author}{\bibinfo{person}{Haoyang Huang}, \bibinfo{person}{Yaobo
  Liang}, \bibinfo{person}{Nan Duan}, \bibinfo{person}{Ming Gong},
  \bibinfo{person}{Linjun Shou}, \bibinfo{person}{Daxin Jiang}, {and}
  \bibinfo{person}{Ming Zhou}.} \bibinfo{year}{2019}\natexlab{}.
\newblock \showarticletitle{{U}nicoder: A Universal Language Encoder by
  Pre-training with Multiple Cross-lingual Tasks}. In
  \bibinfo{booktitle}{\emph{Proceedings of the 2019 Conference on Empirical
  Methods in Natural Language Processing and the 9th International Joint
  Conference on Natural Language Processing (EMNLP-IJCNLP)}}.
  \bibinfo{publisher}{Association for Computational Linguistics},
  \bibinfo{address}{Hong Kong, China}, \bibinfo{pages}{2485--2494}.
\newblock
\urldef\tempurl%
\url{https://doi.org/10.18653/v1/D19-1252}
\showDOI{\tempurl}


\bibitem[\protect\citeauthoryear{Johnson, Schuster, Le, Krikun, Wu, Chen,
  Thorat, Vi{\'e}gas, Wattenberg, Corrado, Hughes, and Dean}{Johnson
  et~al\mbox{.}}{2017}]%
        {johnson-etal-2017-googles}
\bibfield{author}{\bibinfo{person}{Melvin Johnson}, \bibinfo{person}{Mike
  Schuster}, \bibinfo{person}{Quoc~V. Le}, \bibinfo{person}{Maxim Krikun},
  \bibinfo{person}{Yonghui Wu}, \bibinfo{person}{Zhifeng Chen},
  \bibinfo{person}{Nikhil Thorat}, \bibinfo{person}{Fernanda Vi{\'e}gas},
  \bibinfo{person}{Martin Wattenberg}, \bibinfo{person}{Greg Corrado},
  \bibinfo{person}{Macduff Hughes}, {and} \bibinfo{person}{Jeffrey Dean}.}
  \bibinfo{year}{2017}\natexlab{}.
\newblock \showarticletitle{{G}oogle{'}s Multilingual Neural Machine
  Translation System: Enabling Zero-Shot Translation}.
\newblock \bibinfo{journal}{\emph{Transactions of the Association for
  Computational Linguistics}}  \bibinfo{volume}{5} (\bibinfo{year}{2017}),
  \bibinfo{pages}{339--351}.
\newblock
\urldef\tempurl%
\url{https://doi.org/10.1162/tacl_a_00065}
\showDOI{\tempurl}


\bibitem[\protect\citeauthoryear{Joshi, Barnes, Santy, Khanuja, Shah,
  Srinivasan, Bhattamishra, Sitaram, Choudhury, and Bali}{Joshi
  et~al\mbox{.}}{2019}]%
        {joshi-etal-2019-unsung}
\bibfield{author}{\bibinfo{person}{Pratik Joshi}, \bibinfo{person}{Christain
  Barnes}, \bibinfo{person}{Sebastin Santy}, \bibinfo{person}{Simran Khanuja},
  \bibinfo{person}{Sanket Shah}, \bibinfo{person}{Anirudh Srinivasan},
  \bibinfo{person}{Satwik Bhattamishra}, \bibinfo{person}{Sunayana Sitaram},
  \bibinfo{person}{Monojit Choudhury}, {and} \bibinfo{person}{Kalika Bali}.}
  \bibinfo{year}{2019}\natexlab{}.
\newblock \showarticletitle{Unsung Challenges of Building and Deploying
  Language Technologies for Low Resource Language Communities}. In
  \bibinfo{booktitle}{\emph{Proceedings of the 16th International Conference on
  Natural Language Processing}}. \bibinfo{publisher}{NLP Association of India},
  \bibinfo{address}{International Institute of Information Technology,
  Hyderabad, India}, \bibinfo{pages}{211--219}.
\newblock
\urldef\tempurl%
\url{https://aclanthology.org/2019.icon-1.25}
\showURL{%
\tempurl}


\bibitem[\protect\citeauthoryear{Joshi, Santy, Budhiraja, Bali, and
  Choudhury}{Joshi et~al\mbox{.}}{2020}]%
        {joshi-etal-2020-state}
\bibfield{author}{\bibinfo{person}{Pratik Joshi}, \bibinfo{person}{Sebastin
  Santy}, \bibinfo{person}{Amar Budhiraja}, \bibinfo{person}{Kalika Bali},
  {and} \bibinfo{person}{Monojit Choudhury}.} \bibinfo{year}{2020}\natexlab{}.
\newblock \showarticletitle{The State and Fate of Linguistic Diversity and
  Inclusion in the {NLP} World}. In \bibinfo{booktitle}{\emph{Proceedings of
  the 58th Annual Meeting of the Association for Computational Linguistics}}.
  \bibinfo{publisher}{Association for Computational Linguistics},
  \bibinfo{address}{Online}, \bibinfo{pages}{6282--6293}.
\newblock
\urldef\tempurl%
\url{https://doi.org/10.18653/v1/2020.acl-main.560}
\showDOI{\tempurl}


\bibitem[\protect\citeauthoryear{K, Wang, Mayhew, and Roth}{K
  et~al\mbox{.}}{2020}]%
        {k2020crosslingual}
\bibfield{author}{\bibinfo{person}{Karthikeyan K}, \bibinfo{person}{Zihan
  Wang}, \bibinfo{person}{Stephen Mayhew}, {and} \bibinfo{person}{Dan Roth}.}
  \bibinfo{year}{2020}\natexlab{}.
\newblock \bibinfo{title}{Cross-Lingual Ability of Multilingual BERT: An
  Empirical Study}.
\newblock
\newblock
\showeprint[arxiv]{1912.07840}~[cs.CL]


\bibitem[\protect\citeauthoryear{Khanuja, Dandapat, Srinivasan, Sitaram, and
  Choudhury}{Khanuja et~al\mbox{.}}{2020}]%
        {khanuja-etal-2020-gluecos}
\bibfield{author}{\bibinfo{person}{Simran Khanuja}, \bibinfo{person}{Sandipan
  Dandapat}, \bibinfo{person}{Anirudh Srinivasan}, \bibinfo{person}{Sunayana
  Sitaram}, {and} \bibinfo{person}{Monojit Choudhury}.}
  \bibinfo{year}{2020}\natexlab{}.
\newblock \showarticletitle{{GLUEC}o{S}: An Evaluation Benchmark for
  Code-Switched {NLP}}. In \bibinfo{booktitle}{\emph{Proceedings of the 58th
  Annual Meeting of the Association for Computational Linguistics}}.
  \bibinfo{publisher}{Association for Computational Linguistics},
  \bibinfo{address}{Online}, \bibinfo{pages}{3575--3585}.
\newblock
\urldef\tempurl%
\url{https://doi.org/10.18653/v1/2020.acl-main.329}
\showDOI{\tempurl}


\bibitem[\protect\citeauthoryear{Lample and Conneau}{Lample and
  Conneau}{2019}]%
        {lample2019crosslingual}
\bibfield{author}{\bibinfo{person}{Guillaume Lample} {and}
  \bibinfo{person}{Alexis Conneau}.} \bibinfo{year}{2019}\natexlab{}.
\newblock \bibinfo{title}{Cross-lingual Language Model Pretraining}.
\newblock
\newblock
\showeprint[arxiv]{1901.07291}~[cs.CL]


\bibitem[\protect\citeauthoryear{Lauscher, Ravishankar, Vuli{\'c}, and
  Glava{\v{s}}}{Lauscher et~al\mbox{.}}{2020}]%
        {lauscher-etal-2020-zero}
\bibfield{author}{\bibinfo{person}{Anne Lauscher}, \bibinfo{person}{Vinit
  Ravishankar}, \bibinfo{person}{Ivan Vuli{\'c}}, {and} \bibinfo{person}{Goran
  Glava{\v{s}}}.} \bibinfo{year}{2020}\natexlab{}.
\newblock \showarticletitle{From Zero to Hero: {O}n the Limitations of
  Zero-Shot Language Transfer with Multilingual {T}ransformers}. In
  \bibinfo{booktitle}{\emph{Proceedings of the 2020 Conference on Empirical
  Methods in Natural Language Processing (EMNLP)}}.
  \bibinfo{publisher}{Association for Computational Linguistics},
  \bibinfo{address}{Online}, \bibinfo{pages}{4483--4499}.
\newblock
\urldef\tempurl%
\url{https://doi.org/10.18653/v1/2020.emnlp-main.363}
\showDOI{\tempurl}


\bibitem[\protect\citeauthoryear{Lewis, Oguz, Rinott, Riedel, and
  Schwenk}{Lewis et~al\mbox{.}}{2020}]%
        {lewis-etal-2020-mlqa}
\bibfield{author}{\bibinfo{person}{Patrick Lewis}, \bibinfo{person}{Barlas
  Oguz}, \bibinfo{person}{Ruty Rinott}, \bibinfo{person}{Sebastian Riedel},
  {and} \bibinfo{person}{Holger Schwenk}.} \bibinfo{year}{2020}\natexlab{}.
\newblock \showarticletitle{{MLQA}: Evaluating Cross-lingual Extractive
  Question Answering}. In \bibinfo{booktitle}{\emph{Proceedings of the 58th
  Annual Meeting of the Association for Computational Linguistics}}.
  \bibinfo{publisher}{Association for Computational Linguistics},
  \bibinfo{address}{Online}, \bibinfo{pages}{7315--7330}.
\newblock
\urldef\tempurl%
\url{https://doi.org/10.18653/v1/2020.acl-main.653}
\showDOI{\tempurl}


\bibitem[\protect\citeauthoryear{Lin, Chen, Lee, Li, Zhang, Xia, Rijhwani, He,
  Zhang, Ma, Anastasopoulos, Littell, and Neubig}{Lin et~al\mbox{.}}{2019}]%
        {lin-etal-2019-choosing}
\bibfield{author}{\bibinfo{person}{Yu-Hsiang Lin}, \bibinfo{person}{Chian-Yu
  Chen}, \bibinfo{person}{Jean Lee}, \bibinfo{person}{Zirui Li},
  \bibinfo{person}{Yuyan Zhang}, \bibinfo{person}{Mengzhou Xia},
  \bibinfo{person}{Shruti Rijhwani}, \bibinfo{person}{Junxian He},
  \bibinfo{person}{Zhisong Zhang}, \bibinfo{person}{Xuezhe Ma},
  \bibinfo{person}{Antonios Anastasopoulos}, \bibinfo{person}{Patrick Littell},
  {and} \bibinfo{person}{Graham Neubig}.} \bibinfo{year}{2019}\natexlab{}.
\newblock \showarticletitle{Choosing Transfer Languages for Cross-Lingual
  Learning}. In \bibinfo{booktitle}{\emph{Proceedings of the 57th Annual
  Meeting of the Association for Computational Linguistics}}.
  \bibinfo{publisher}{Association for Computational Linguistics},
  \bibinfo{address}{Florence, Italy}, \bibinfo{pages}{3125--3135}.
\newblock
\urldef\tempurl%
\url{https://doi.org/10.18653/v1/P19-1301}
\showDOI{\tempurl}


\bibitem[\protect\citeauthoryear{Littell, Mortensen, Lin, Kairis, Turner, and
  Levin}{Littell et~al\mbox{.}}{2017}]%
        {littell-etal-2017-uriel}
\bibfield{author}{\bibinfo{person}{Patrick Littell}, \bibinfo{person}{David~R.
  Mortensen}, \bibinfo{person}{Ke Lin}, \bibinfo{person}{Katherine Kairis},
  \bibinfo{person}{Carlisle Turner}, {and} \bibinfo{person}{Lori Levin}.}
  \bibinfo{year}{2017}\natexlab{}.
\newblock \showarticletitle{{URIEL} and lang2vec: Representing languages as
  typological, geographical, and phylogenetic vectors}. In
  \bibinfo{booktitle}{\emph{Proceedings of the 15th Conference of the
  {E}uropean Chapter of the Association for Computational Linguistics: Volume
  2, Short Papers}}. \bibinfo{publisher}{Association for Computational
  Linguistics}, \bibinfo{address}{Valencia, Spain}, \bibinfo{pages}{8--14}.
\newblock
\urldef\tempurl%
\url{https://www.aclweb.org/anthology/E17-2002}
\showURL{%
\tempurl}


\bibitem[\protect\citeauthoryear{Nguyen and Chiang}{Nguyen and Chiang}{2017}]%
        {nguyen-chiang-2017-transfer}
\bibfield{author}{\bibinfo{person}{Toan~Q. Nguyen} {and} \bibinfo{person}{David
  Chiang}.} \bibinfo{year}{2017}\natexlab{}.
\newblock \showarticletitle{Transfer Learning across Low-Resource, Related
  Languages for Neural Machine Translation}. In
  \bibinfo{booktitle}{\emph{Proceedings of the Eighth International Joint
  Conference on Natural Language Processing (Volume 2: Short Papers)}}.
  \bibinfo{publisher}{Asian Federation of Natural Language Processing},
  \bibinfo{address}{Taipei, Taiwan}, \bibinfo{pages}{296--301}.
\newblock
\urldef\tempurl%
\url{https://www.aclweb.org/anthology/I17-2050}
\showURL{%
\tempurl}


\bibitem[\protect\citeauthoryear{Ortiz~Su{\'a}rez, Romary, and
  Sagot}{Ortiz~Su{\'a}rez et~al\mbox{.}}{2020}]%
        {ortiz-suarez-etal-2020-monolingual}
\bibfield{author}{\bibinfo{person}{Pedro~Javier Ortiz~Su{\'a}rez},
  \bibinfo{person}{Laurent Romary}, {and} \bibinfo{person}{Beno{\^\i}t Sagot}.}
  \bibinfo{year}{2020}\natexlab{}.
\newblock \showarticletitle{A Monolingual Approach to Contextualized Word
  Embeddings for Mid-Resource Languages}. In
  \bibinfo{booktitle}{\emph{Proceedings of the 58th Annual Meeting of the
  Association for Computational Linguistics}}. \bibinfo{publisher}{Association
  for Computational Linguistics}, \bibinfo{address}{Online},
  \bibinfo{pages}{1703--1714}.
\newblock
\urldef\tempurl%
\url{https://doi.org/10.18653/v1/2020.acl-main.156}
\showDOI{\tempurl}


\bibitem[\protect\citeauthoryear{Pires, Schlinger, and Garrette}{Pires
  et~al\mbox{.}}{2019}]%
        {pires-etal-2019-multilingual}
\bibfield{author}{\bibinfo{person}{Telmo Pires}, \bibinfo{person}{Eva
  Schlinger}, {and} \bibinfo{person}{Dan Garrette}.}
  \bibinfo{year}{2019}\natexlab{}.
\newblock \showarticletitle{How Multilingual is Multilingual {BERT}?}. In
  \bibinfo{booktitle}{\emph{Proceedings of the 57th Annual Meeting of the
  Association for Computational Linguistics}}. \bibinfo{publisher}{Association
  for Computational Linguistics}, \bibinfo{address}{Florence, Italy},
  \bibinfo{pages}{4996--5001}.
\newblock
\urldef\tempurl%
\url{https://doi.org/10.18653/v1/P19-1493}
\showDOI{\tempurl}


\bibitem[\protect\citeauthoryear{Pyysalo, Kanerva, Virtanen, and
  Ginter}{Pyysalo et~al\mbox{.}}{2021}]%
        {pyysalo-etal-2021-wikibert}
\bibfield{author}{\bibinfo{person}{Sampo Pyysalo}, \bibinfo{person}{Jenna
  Kanerva}, \bibinfo{person}{Antti Virtanen}, {and} \bibinfo{person}{Filip
  Ginter}.} \bibinfo{year}{2021}\natexlab{}.
\newblock \showarticletitle{{W}iki{BERT} Models: Deep Transfer Learning for
  Many Languages}. In \bibinfo{booktitle}{\emph{Proceedings of the 23rd Nordic
  Conference on Computational Linguistics (NoDaLiDa)}}.
  \bibinfo{publisher}{Link{\"o}ping University Electronic Press, Sweden},
  \bibinfo{address}{Reykjavik, Iceland (Online)}, \bibinfo{pages}{1--10}.
\newblock
\urldef\tempurl%
\url{https://aclanthology.org/2021.nodalida-main.1}
\showURL{%
\tempurl}


\bibitem[\protect\citeauthoryear{Ribeiro, Wu, Guestrin, and Singh}{Ribeiro
  et~al\mbox{.}}{2020}]%
        {ribeiro-etal-2020-beyond}
\bibfield{author}{\bibinfo{person}{Marco~Tulio Ribeiro},
  \bibinfo{person}{Tongshuang Wu}, \bibinfo{person}{Carlos Guestrin}, {and}
  \bibinfo{person}{Sameer Singh}.} \bibinfo{year}{2020}\natexlab{}.
\newblock \showarticletitle{Beyond Accuracy: Behavioral Testing of {NLP} Models
  with {C}heck{L}ist}. In \bibinfo{booktitle}{\emph{Proceedings of the 58th
  Annual Meeting of the Association for Computational Linguistics}}.
  \bibinfo{publisher}{Association for Computational Linguistics},
  \bibinfo{address}{Online}, \bibinfo{pages}{4902--4912}.
\newblock
\urldef\tempurl%
\url{https://doi.org/10.18653/v1/2020.acl-main.442}
\showDOI{\tempurl}


\bibitem[\protect\citeauthoryear{Tjong Kim~Sang and De~Meulder}{Tjong Kim~Sang
  and De~Meulder}{2003}]%
        {tjong-kim-sang-de-meulder-2003-introduction}
\bibfield{author}{\bibinfo{person}{Erik~F. Tjong Kim~Sang} {and}
  \bibinfo{person}{Fien De~Meulder}.} \bibinfo{year}{2003}\natexlab{}.
\newblock \showarticletitle{Introduction to the {C}o{NLL}-2003 Shared Task:
  Language-Independent Named Entity Recognition}. In
  \bibinfo{booktitle}{\emph{Proceedings of the Seventh Conference on Natural
  Language Learning at {HLT}-{NAACL} 2003}}. \bibinfo{pages}{142--147}.
\newblock
\urldef\tempurl%
\url{https://www.aclweb.org/anthology/W03-0419}
\showURL{%
\tempurl}


\bibitem[\protect\citeauthoryear{Turc, Lee, Eisenstein, Chang, and
  Toutanova}{Turc et~al\mbox{.}}{2021}]%
        {turc2021revisiting}
\bibfield{author}{\bibinfo{person}{Iulia Turc}, \bibinfo{person}{Kenton Lee},
  \bibinfo{person}{Jacob Eisenstein}, \bibinfo{person}{Ming-Wei Chang}, {and}
  \bibinfo{person}{Kristina Toutanova}.} \bibinfo{year}{2021}\natexlab{}.
\newblock \bibinfo{title}{Revisiting the Primacy of English in Zero-shot
  Cross-lingual Transfer}.
\newblock
\newblock
\showeprint[arxiv]{2106.16171}~[cs.CL]


\bibitem[\protect\citeauthoryear{Virtanen, Kanerva, Ilo, Luoma, Luotolahti,
  Salakoski, Ginter, and Pyysalo}{Virtanen et~al\mbox{.}}{2019}]%
        {virtanen2019multilingual}
\bibfield{author}{\bibinfo{person}{Antti Virtanen}, \bibinfo{person}{Jenna
  Kanerva}, \bibinfo{person}{Rami Ilo}, \bibinfo{person}{Jouni Luoma},
  \bibinfo{person}{Juhani Luotolahti}, \bibinfo{person}{Tapio Salakoski},
  \bibinfo{person}{Filip Ginter}, {and} \bibinfo{person}{Sampo Pyysalo}.}
  \bibinfo{year}{2019}\natexlab{}.
\newblock \bibinfo{title}{Multilingual is not enough: BERT for Finnish}.
\newblock
\newblock
\showeprint[arxiv]{1912.07076}~[cs.CL]


\bibitem[\protect\citeauthoryear{Vu, Wang, Munkhdalai, Sordoni, Trischler,
  Mattarella-Micke, Maji, and Iyyer}{Vu et~al\mbox{.}}{2020}]%
        {vu-etal-2020-exploring}
\bibfield{author}{\bibinfo{person}{Tu Vu}, \bibinfo{person}{Tong Wang},
  \bibinfo{person}{Tsendsuren Munkhdalai}, \bibinfo{person}{Alessandro
  Sordoni}, \bibinfo{person}{Adam Trischler}, \bibinfo{person}{Andrew
  Mattarella-Micke}, \bibinfo{person}{Subhransu Maji}, {and}
  \bibinfo{person}{Mohit Iyyer}.} \bibinfo{year}{2020}\natexlab{}.
\newblock \showarticletitle{Exploring and Predicting Transferability across
  {NLP} Tasks}. In \bibinfo{booktitle}{\emph{Proceedings of the 2020 Conference
  on Empirical Methods in Natural Language Processing (EMNLP)}}.
  \bibinfo{publisher}{Association for Computational Linguistics},
  \bibinfo{address}{Online}, \bibinfo{pages}{7882--7926}.
\newblock
\urldef\tempurl%
\url{https://doi.org/10.18653/v1/2020.emnlp-main.635}
\showDOI{\tempurl}


\bibitem[\protect\citeauthoryear{Wu and Dredze}{Wu and Dredze}{2019}]%
        {wu-dredze-2019-beto}
\bibfield{author}{\bibinfo{person}{Shijie Wu} {and} \bibinfo{person}{Mark
  Dredze}.} \bibinfo{year}{2019}\natexlab{}.
\newblock \showarticletitle{Beto, Bentz, Becas: The Surprising Cross-Lingual
  Effectiveness of {BERT}}. In \bibinfo{booktitle}{\emph{Proceedings of the
  2019 Conference on Empirical Methods in Natural Language Processing and the
  9th International Joint Conference on Natural Language Processing
  (EMNLP-IJCNLP)}}. \bibinfo{publisher}{Association for Computational
  Linguistics}, \bibinfo{address}{Hong Kong, China}, \bibinfo{pages}{833--844}.
\newblock
\urldef\tempurl%
\url{https://doi.org/10.18653/v1/D19-1077}
\showDOI{\tempurl}


\bibitem[\protect\citeauthoryear{Wu and Dredze}{Wu and Dredze}{2020}]%
        {wu-dredze-2020-languages}
\bibfield{author}{\bibinfo{person}{Shijie Wu} {and} \bibinfo{person}{Mark
  Dredze}.} \bibinfo{year}{2020}\natexlab{}.
\newblock \showarticletitle{Are All Languages Created Equal in Multilingual
  {BERT}?}. In \bibinfo{booktitle}{\emph{Proceedings of the 5th Workshop on
  Representation Learning for NLP}}. \bibinfo{publisher}{Association for
  Computational Linguistics}, \bibinfo{address}{Online},
  \bibinfo{pages}{120--130}.
\newblock
\urldef\tempurl%
\url{https://doi.org/10.18653/v1/2020.repl4nlp-1.16}
\showDOI{\tempurl}


\bibitem[\protect\citeauthoryear{Xia, Anastasopoulos, Xu, Yang, and Neubig}{Xia
  et~al\mbox{.}}{2020}]%
        {xia-etal-2020-predicting}
\bibfield{author}{\bibinfo{person}{Mengzhou Xia}, \bibinfo{person}{Antonios
  Anastasopoulos}, \bibinfo{person}{Ruochen Xu}, \bibinfo{person}{Yiming Yang},
  {and} \bibinfo{person}{Graham Neubig}.} \bibinfo{year}{2020}\natexlab{}.
\newblock \showarticletitle{Predicting Performance for Natural Language
  Processing Tasks}. In \bibinfo{booktitle}{\emph{Proceedings of the 58th
  Annual Meeting of the Association for Computational Linguistics}}.
  \bibinfo{publisher}{Association for Computational Linguistics},
  \bibinfo{address}{Online}, \bibinfo{pages}{8625--8646}.
\newblock
\urldef\tempurl%
\url{https://doi.org/10.18653/v1/2020.acl-main.764}
\showDOI{\tempurl}


\bibitem[\protect\citeauthoryear{Xue, Constant, Roberts, Kale, Al-Rfou,
  Siddhant, Barua, and Raffel}{Xue et~al\mbox{.}}{2021}]%
        {xue2021mt5}
\bibfield{author}{\bibinfo{person}{Linting Xue}, \bibinfo{person}{Noah
  Constant}, \bibinfo{person}{Adam Roberts}, \bibinfo{person}{Mihir Kale},
  \bibinfo{person}{Rami Al-Rfou}, \bibinfo{person}{Aditya Siddhant},
  \bibinfo{person}{Aditya Barua}, {and} \bibinfo{person}{Colin Raffel}.}
  \bibinfo{year}{2021}\natexlab{}.
\newblock \bibinfo{title}{mT5: A massively multilingual pre-trained
  text-to-text transformer}.
\newblock
\newblock
\showeprint[arxiv]{2010.11934}~[cs.CL]


\bibitem[\protect\citeauthoryear{Yang, Zhang, Tar, and Baldridge}{Yang
  et~al\mbox{.}}{2019}]%
        {yang-etal-2019-paws}
\bibfield{author}{\bibinfo{person}{Yinfei Yang}, \bibinfo{person}{Yuan Zhang},
  \bibinfo{person}{Chris Tar}, {and} \bibinfo{person}{Jason Baldridge}.}
  \bibinfo{year}{2019}\natexlab{}.
\newblock \showarticletitle{{PAWS}-{X}: A Cross-lingual Adversarial Dataset for
  Paraphrase Identification}. In \bibinfo{booktitle}{\emph{Proceedings of the
  2019 Conference on Empirical Methods in Natural Language Processing and the
  9th International Joint Conference on Natural Language Processing
  (EMNLP-IJCNLP)}}. \bibinfo{publisher}{Association for Computational
  Linguistics}, \bibinfo{address}{Hong Kong, China},
  \bibinfo{pages}{3687--3692}.
\newblock
\urldef\tempurl%
\url{https://doi.org/10.18653/v1/D19-1382}
\showDOI{\tempurl}


\bibitem[\protect\citeauthoryear{Ye, Liu, Fu, and Neubig}{Ye
  et~al\mbox{.}}{2021}]%
        {ye-etal-2021-towards}
\bibfield{author}{\bibinfo{person}{Zihuiwen Ye}, \bibinfo{person}{Pengfei Liu},
  \bibinfo{person}{Jinlan Fu}, {and} \bibinfo{person}{Graham Neubig}.}
  \bibinfo{year}{2021}\natexlab{}.
\newblock \showarticletitle{Towards More Fine-grained and Reliable {NLP}
  Performance Prediction}. In \bibinfo{booktitle}{\emph{Proceedings of the 16th
  Conference of the European Chapter of the Association for Computational
  Linguistics: Main Volume}}. \bibinfo{publisher}{Association for Computational
  Linguistics}, \bibinfo{address}{Online}, \bibinfo{pages}{3703--3714}.
\newblock
\urldef\tempurl%
\url{https://aclanthology.org/2021.eacl-main.324}
\showURL{%
\tempurl}


\bibitem[\protect\citeauthoryear{Zipf}{Zipf}{1949}]%
        {zipf1949human}
\bibfield{author}{\bibinfo{person}{George~Kingsley Zipf}.}
  \bibinfo{year}{1949}\natexlab{}.
\newblock \showarticletitle{Human behavior and the principle of least effort.}
\newblock  (\bibinfo{year}{1949}).
\newblock


\end{thebibliography}

\appendix

\end{document}